\documentclass[10pt,twocolumn,letterpaper]{article}

\usepackage{cvpr}
\usepackage{times}
\usepackage{epsfig}
\usepackage{graphicx}
\usepackage{amsmath}
\usepackage{amssymb}
\usepackage{float}

\usepackage[breaklinks=true,bookmarks=false]{hyperref}

\cvprfinalcopy 

\def\cvprPaperID{3510} 

\newcommand{\bR}{\mathbb{R}}

\newcommand{\mat}[1]{\mathbf{#1}}

\newcommand{\cross}[1]{[#1]_{\times}}

\begin{document}

\title{All Graphs Lead to Rome: Learning Geometric and Cycle-Consistent Representations with Graph Convolutional Networks}

\author{Stephen Phillips, Kostas Daniilidis \\
GRASP Laboratory, University of Pennsylvania\\
{\tt\small \{stephi, kostas\}@seas.upenn.edu}
}


\maketitle


\begin{abstract}
    Image feature matching is a fundamental part of many geometric computer vision applications, and using multiple images can improve performance.
    In this work, we formulate multi-image matching as a graph embedding problem then use a Graph Convolutional Network to learn an appropriate embedding function for aligning image features.
    We use cycle consistency to train our network in an unsupervised fashion, since ground truth correspondence is difficult or expensive to aquire.
    In addition, geometric consistency losses can be added at training time, even if the information is not available in the test set, unlike previous approaches that optimize cycle consistency directly.
    To the best of our knowledge, no other works have used learning for multi-image feature matching.
    Our experiments show that our method is competitive with other optimization based approaches.
\end{abstract}

\section{Introduction}

\begin{figure}[t]
\begin{center}
  \includegraphics[width=0.8\linewidth]{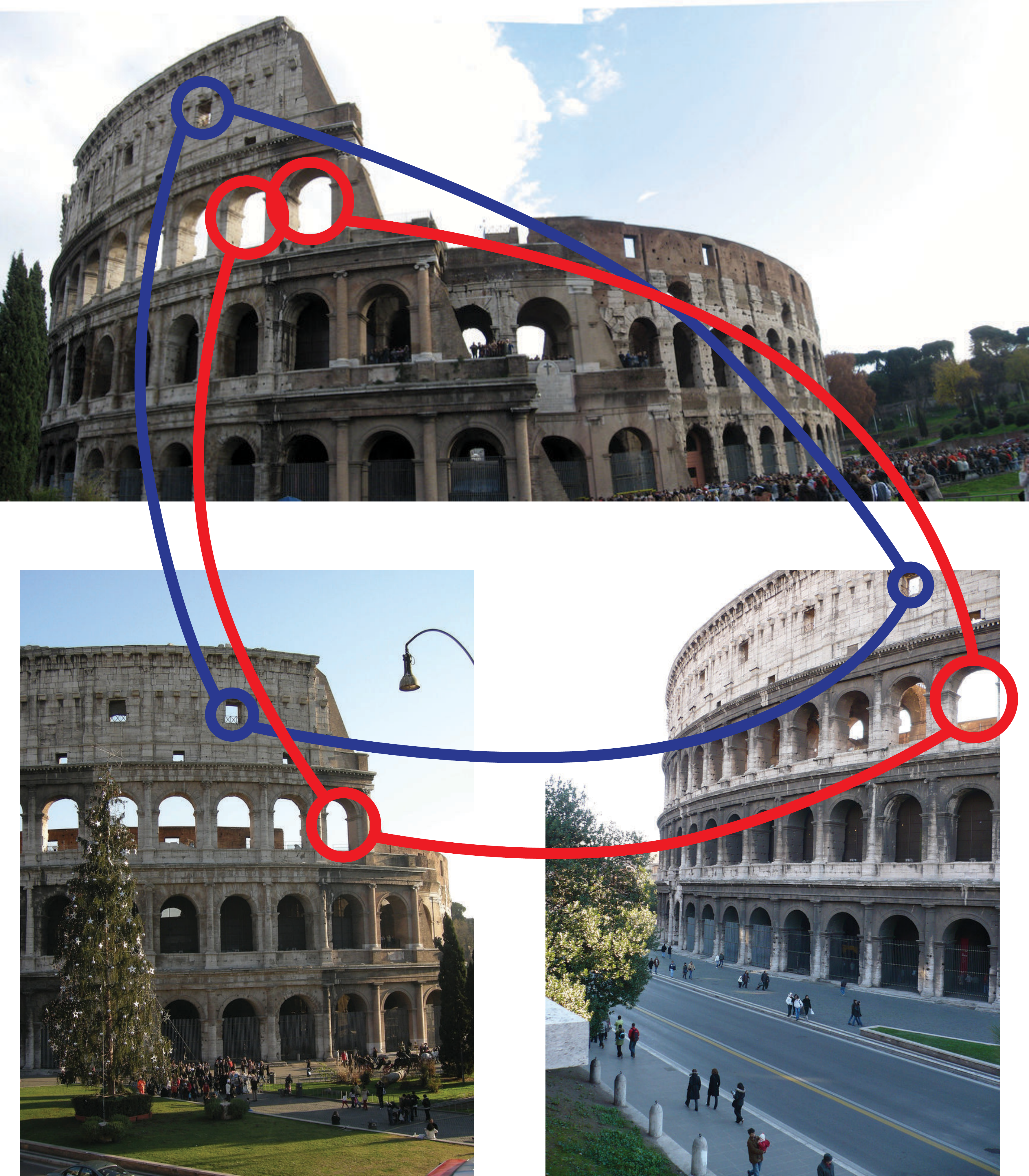}
\end{center}
  \caption{
    Example of multi-image matching and cycle consistency, with images from Rome16K dataset \cite{li2010location}.
    Pairwise matches between image can lead to global inconsistencies in the matching.
    In this example, an error in matching leads to an inconsistent cycle (shown in red), ideally these pairwise matches would be consistent.
    We improve the quality of pairwise matches in the multi-view matching problem by training a neural net with cycle consistency.
  }
\label{fig:cycconsistex}
\label{fig:onecol}
\end{figure}

Feature matching is an essential part of Structure from Motion and many geometric computer vision applications.
The goal in multi-image feature matching is to take 2D image locations from three or more images and find which ones correspond to the same point in the 3D scene.
Methods such as SIFT feature matching \cite{lowe2004distinctive} followed by RANSAC \cite{fischler1981random} have been the standard for decades.
However RANSAC-based approaches are limited to matching pairs of images, which can lead to global inconsistencies in the matching.
Other works, such as Wang et al.~\cite{wang2017multi}, have shown improvement in performance by optimizing cycle consistency, i.e. enforcing the pairwise feature matches to be globally consistent, illustrated in figure \ref{fig:cycconsistex}.

However, these multi-view consistency algorithms struggle in distributed and robust settings.
Deep learning has revolutionized how image features are computed \cite{yi2016lift}.
In this paper, we want to leverage the power of deep representations in order to compute feature descriptors that are robust across multiple views.

\begin{figure*}[t]
\begin{center}
  \includegraphics[width=0.8\linewidth]{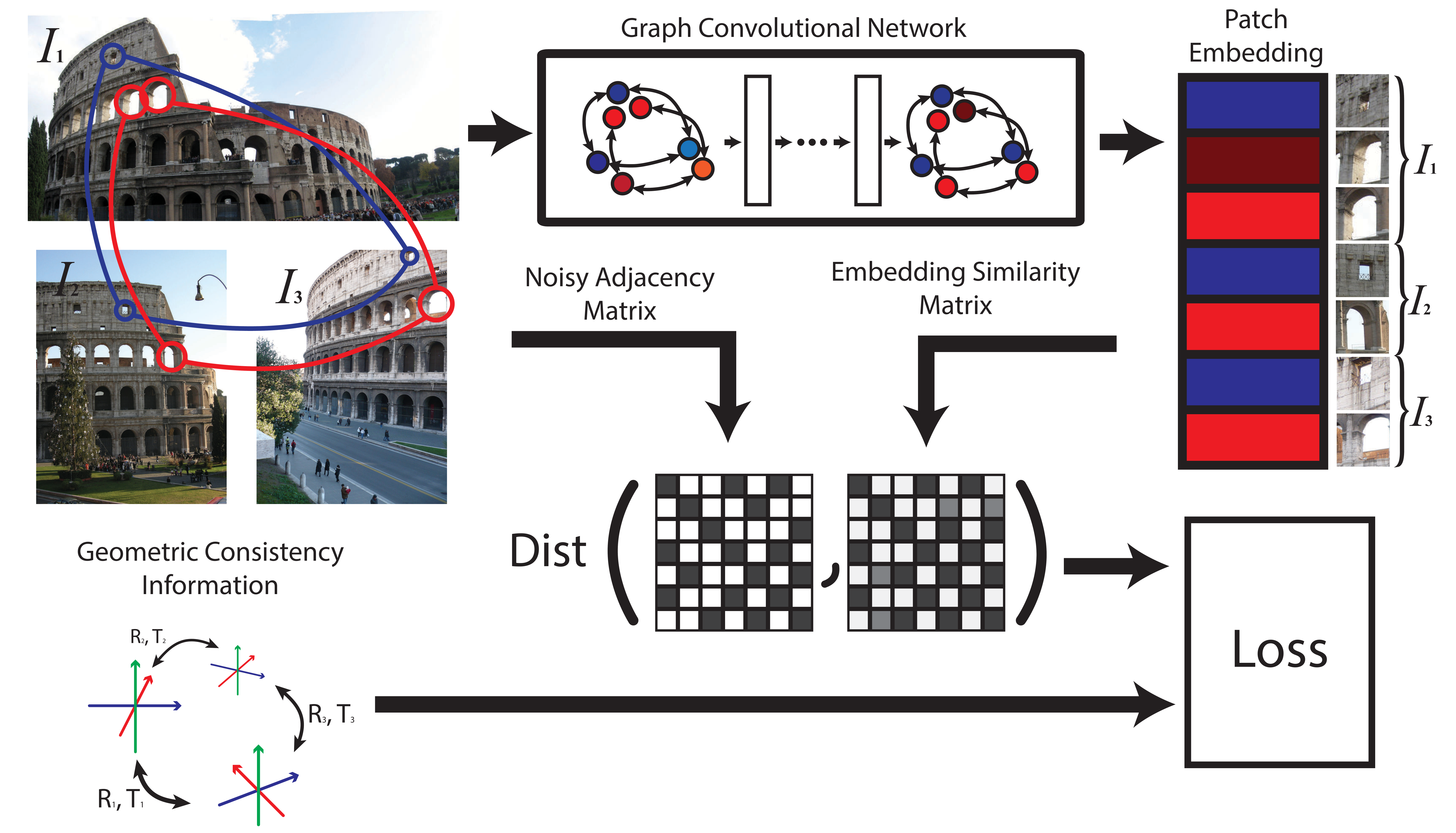}
\end{center}
  \caption{
    An illustration of the approach of this work.
    The Graph Convolutional Neural Network (GCN) \cite{kipf2016semi} takes as input the graph of matches and then outputs a low rank embedding of the adjacency matrix of the graph.
    The GCN operates on an embedding over the nodes of the graph.
    In the figure, the GCN node embeddings are represented by different colors.
    The final embedding is used to construct a pairwise similarity matrix, which we train to be a low dimensional cycle-consistent representation of the graph adjacency matrix, thus pruning the erroneous matches.
    We train the network using a reconstruction loss on the similarity matrix with the noisy adjacency matrix, and thus do not need ground truth matches.
    In addition, we can use geometric consistency information, such as epipolar constraints, to assist training the network.
  }
\label{fig:pipeline}
\label{fig:onecol}
\end{figure*}

Unfortunately, there are obstacles to applying multi-view constraints directly to deep learning. 
To train networks, we need a large amount of labeled data.
In the case of multi-image feature matching, one would need hand labeled point correspondences between images, which are difficult and expensive to obtain.
Multi-view constraints are formulated in terms of sparse features, which traditional convolutional neural nets are not designed to handle.
Additionally, in the case of multi-image feature matching, geometric constraints would be helpful in rejecting outlier matches.

In this work, we propose to solve these problems using a Graph Convolutional Network (GCN) to operate on the correspondence graph.
The proposed method works directly on the correspondence graph, which is agnostic to how the correspondences were computed, thus allowing the algorithm to work in a broad class of environments.
To the best of our knowledge this work is the first to apply deep learning to the multi-view feature matching problem.
We use an unsupervised loss - the cycle consistency loss - to train the network, and thus avoiding the difficulty of expensive hand labeling.
Geometric consistency losses can aid training, even if such information is not available at test time.
Although our network is simple, it shows promising results compared to baselines which optimize for cycle-consistency without learned embeddings.
Furthermore, since inference requires only a single forward pass over the neural network, our approach is faster (to achieve comparable accuracy) than methods which must solve an optimization problem every time.
Our contributions are:
\begin{itemize}
\item We use a novel architecture to address the multi-image feature matching problem using GCNs with graph embeddings.
\item We introduce an unsupervised cycle consistency loss that does not require labeled correspondences to train.
\item We demonstrate the effectiveness of geometric consistency losses in improving training.
\item We perform experiments on the Rome16K \cite{li2010location} dataset to test the effectiveness of our method compared to optimization based methods.
\end{itemize}

\section{Related Work}

\subsection{Feature Matching}
Matching has a rich history of research in computer vision.
The most well known and widely used method for image matching is RANSAC \cite{fischler1981random}.
RANSAC, combined with hand-crafted feature descriptors such as SIFT \cite{lowe2004distinctive}, SURF \cite{bay2006surf}, BRIEF \cite{calonder2012brief}, or ORB \cite{mur2015orb}, has constituted the bulk of the matching literature for the last 40 years.
Finally, graph matching \cite{suh2015subgraph, hu2016distributable} can be used as a final step for more robust matches.

\subsection{Multi-image Matching}
Multi-image matching has traditionally been done using optimization based methods minimizing a cycle consistency based loss (see Section 3.3).
Pachauri et al.~\cite{pachauri2013solving} and Arrigoni et al.~\cite{arrigoni2017synchronization} use the eigenvectors of the matching matrix to obtain a low dimensional embedding. 
However, the low Gaussian noise assumption is not realistic.
Zhou et al.~\cite{zhou2015multi} and Wang et al.~\cite{wang2017multi} use most sophisticated optimization techniques on the matching matrix and thus produce more robust solutions.
Leonardos et al.~\cite{leonardos2016distributed} implement a distributed optimization scheme to solve for cycle consistency.
As an alternative to optimization based techniques, Tron et al.~\cite{tron2017fast} used density based clustering techniques to compute multi-image correspondence.
To the best of our knowledge, we are the first to use neural networks for multi-image matching.

\subsection{Deep Learning for Matching}
Previous attempts to improve image matching techniques using machine learning have focused on learning the descriptors given ground truth correspondence from curated datasets \cite{zagoruyko2015learning, yi2016lift, brachmann2017dsac}.
This approach is limited if one do not have the ability to get the ground truth correspondences.
There are other methods to build correspondences such as Choy et al.~\cite{choy2016universal}, but they only handle two-view constraints and require dense correspondences.
Most similar to our work, Yi et al.~\cite{yi2018learning} attempts to improve correspondences by learning match probabilities for RANSAC for greater robustness and speed.
However, they only focus on two view matching and do not exploit the advantages of the correspondence structure.
Note that while Zhu et al.~\cite{zhu2017unpaired} use cycle consistency in their loss, their method is restricted to pairwise cycle consistency, while we use multi-image cycle consistency.

\subsection{Graph Neural Networks}
Graph neural networks have recieved more attention recently \cite{bronstein2017geometric, bruna2013spectral, defferrard2016convolutional, kipf2016semi, scarselli2009graph, gama2018mimo, gama2018convolutional, battaglia2018relational}.
Classical so-called Spectral methods used the eigenvectors of the graph Laplacian to compute convolutions as in Bruna et al.~\cite{bruna2013spectral}, but requires an a-priori known graph structure. 
Non-spectral methods do not require a-priori knowledge \cite{bronstein2017geometric, kipf2016semi, scarselli2009graph, gama2018convolutional}.
Most of these methods use polynomials of the graph Laplacian to compute neighborhood averages.
Gama et al.~2018 \cite{gama2018mimo, gama2018convolutional} formalize this notion and generalize it beyond the use of the graph Laplacian.
To improve performance, more sophisticated aggregation techniques and global information passing can be used as discussed in Battaglia et al.~\cite{battaglia2018relational}.

\section{Method}

\begin{figure}[t]
\begin{center}
  \includegraphics[width=0.9\linewidth]{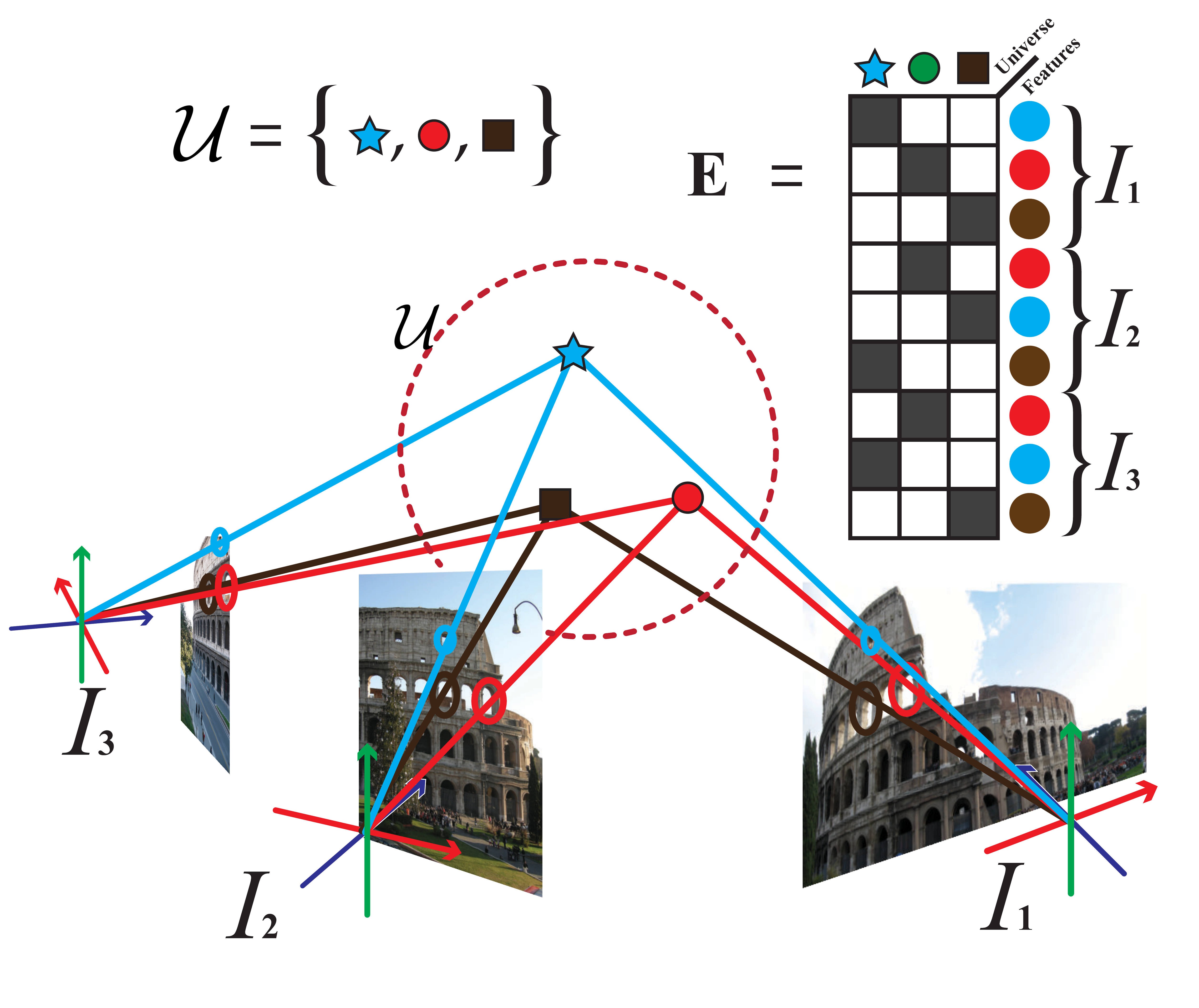}
\end{center}
  \caption{
    Illustration of the universe of features.
    Each feature in each image corresponds to a 3D point in the scene.
    We can construct cycle consistent embeddings of the features by mapping each one to the one-hot vector of its corresponding 3D point.
    While there can be many features, there are fewer 3D points and thus this corresponds to a low rank factorization of the correspondence matrix.
    Best viewed in color.
  }
\label{fig:universefeatures}
\label{fig:onecol}
\end{figure}
Our goal is to learn optimal features that capture multiple image views by filtering out noisy feature matches.
The input to our algorithm is a set of features and noisy correspondences, and the output is a new set of features where the pairwise similarities of these features correspond to the true matches.
We do this by training the new set of feature embeddings to be cycle consistent.
We formulate this problem in terms of the correspondence graph of the features.
Vectors $\mat{x}$ and matrices $\mat{A}$ are denoted with boldface, and the $i^{th}$ row of a matrix is denoted $(\mat{A})_i$.
\subsection{Correspondence Graph}
We assume there is an initial set of feature matches represented as a graph $\mathcal{G} = (\mathcal{V}, \mathcal{E})$.
Each vertex $v$ of the graph is a feature with its associated descriptor $f_v$. 
The graph is represented by its Weighted Adjacency Matrix $\mat{A}(\mathcal{G}) \in \bR^{n \times n}$, where $(\mat{A}(\mathcal{G}))_{ij}$ is the strength of the match between nodes $i$ and $j$.
We introduce the positive diagonal degree matrix $\mat{D}(\mathcal{G}) \in \bR^{n \times n}$, with $\mat{D}(\mathcal{G})_{ii} = \sum_j (\mat{A}(\mathcal{G}))_{ij}$.
For brevity we denote these matrices $\mat{A}$ and $\mat{D}$ respectively.
We introduce the notion of an embedding matrix $\mat{E}_0$ by concatenating the features $f_v$ from each of the vertices.
We use $\mat{E}_0$ as the initialization for our learning algorithm.
We add any additional knowledge we have into the embedding as well (e.g. scale, orientation, etc.).
Putative correspondences are computed from these features and represented by weighted edges in which the weight gives the strength of the match.
While there are many interesting methods for computing these putative correspondences \cite{suh2015subgraph, yi2018learning}, we do not explore them in this work.

In the absence of noise or outliers, this graph would have a connected component for each visible point in the world.
These components should be mutually disjoint. 
In other words each feature would only have edges to other features corresponding to the same point.
Since features in this case represent unique locations in the scene, no points in the same image would have edges between them.
In the noisy case we expect this structure to be corrupted and thus we need to prune the erroneous edges.

To compute this matching graph, we do pairwise feature matching between the images, creating putative correspondences for each of the features.
Typically these putative correspondences are matched probabilistically, meaning a feature in one image matches to many features in another.
The ambiguity in the matches could come from repeated structures in the scene, insufficiently informative low-level feature descriptors, or just an error in the matching algorithm.
Filtering out these noisy matches is our primary learning goal.

However, this graph structure does not have a regular grid structure and thus we cannot use standard convolutional nets to learn features for this task.
Instead we use graph convolutions to learn feature representations on this space.
We describe our approach in the next section.

\subsection{Graph Neural Networks}
As input to our method we are given the graph Laplacian $\mat{L} \in \bR^{n \times n}$ to encode the graph structure:
\begin{equation}
  \mat{L} =\; \mat{I} - \mat{D}^{-1/2} \mat{A} \mat{D}^{-1/2} \\
\end{equation}
and an initial embedding $\mat{E}_0 \in \bR^{n \times m_0}$.
Many older methods assume $\mat{L}$ is known a-priori \cite{bruna2013spectral}, and can encode graph convolutions using the eigenvectors of $L$.
We do not have this luxury, as the correspondence structure changes from image to image, and thus we use non-spectral Graph Neural Networks.
The standard form of non-spectral Graph Neural Networks is:
\begin{equation}
\mat{E}_{i+1} =\; P\left(\sigma\left( \sum_k \mat{L}^k \mat{E}_i \mat{W}_{i,k} \right)\right)
\end{equation}
Here $\sigma$ is the non-linearity (typically a ReLU), $\mat{W}_{k,i}$ are the learned weights of layer $i$,  and $P$ is the `pooling' operation, known as graph coarsening \cite{bronstein2017geometric, gama2018mimo}.
The maximum degree of $\mat{L}$ determines the number of hops away from a node one layer can access.
In the feature matching problem, we require an embedding for each vertex.
Therefore we cannot perform graph coarsening, and thus only perform the non-linearity step.
This work uses Graph Convolutional Networks, proposed in Kipf and Welling \cite{kipf2016semi}:
\begin{align}
      \widetilde{\mat{L}} =&\; (\mat{D} + \mat{I})^{-1/2} (\mat{A} + \mat{I}) (\mat{D} + \mat{I})^{-1/2} \\
\mat{E}_{l+1} =&\; \sigma\left(\widetilde{\mat{L}} \mat{E}_i \mat{W}_i \right)  \label{eq:graph_conv}
\end{align}

The matrix $\widetilde{\mat{L}}$ is analogous to the graph Laplacian (with better numerical stability properties).
$\widetilde{\mat{L}}$ encodes the structure of the graph and is used to perform the actual graph convolution.
Our network has many layers, and thus the receptive field of the final embedding is quite large, even without the pooling operations.
In principle we could use higher hop neighborhoods \cite{gama2018convolutional}, or more complicated aggregation structures \cite{battaglia2018relational}, but in this work we restrict ourselves to simpler architectures, and still obtain promising results for this task.
For faster training, we use group normalization \cite{wu2018group} and skip connections.
The final output $\mat{E}_n$ gives us a new descriptor for every node.
In the next sections we will discuss the loss functions to train this network.

\begin{figure}[t]
\begin{center}
  \includegraphics[width=0.8\linewidth]{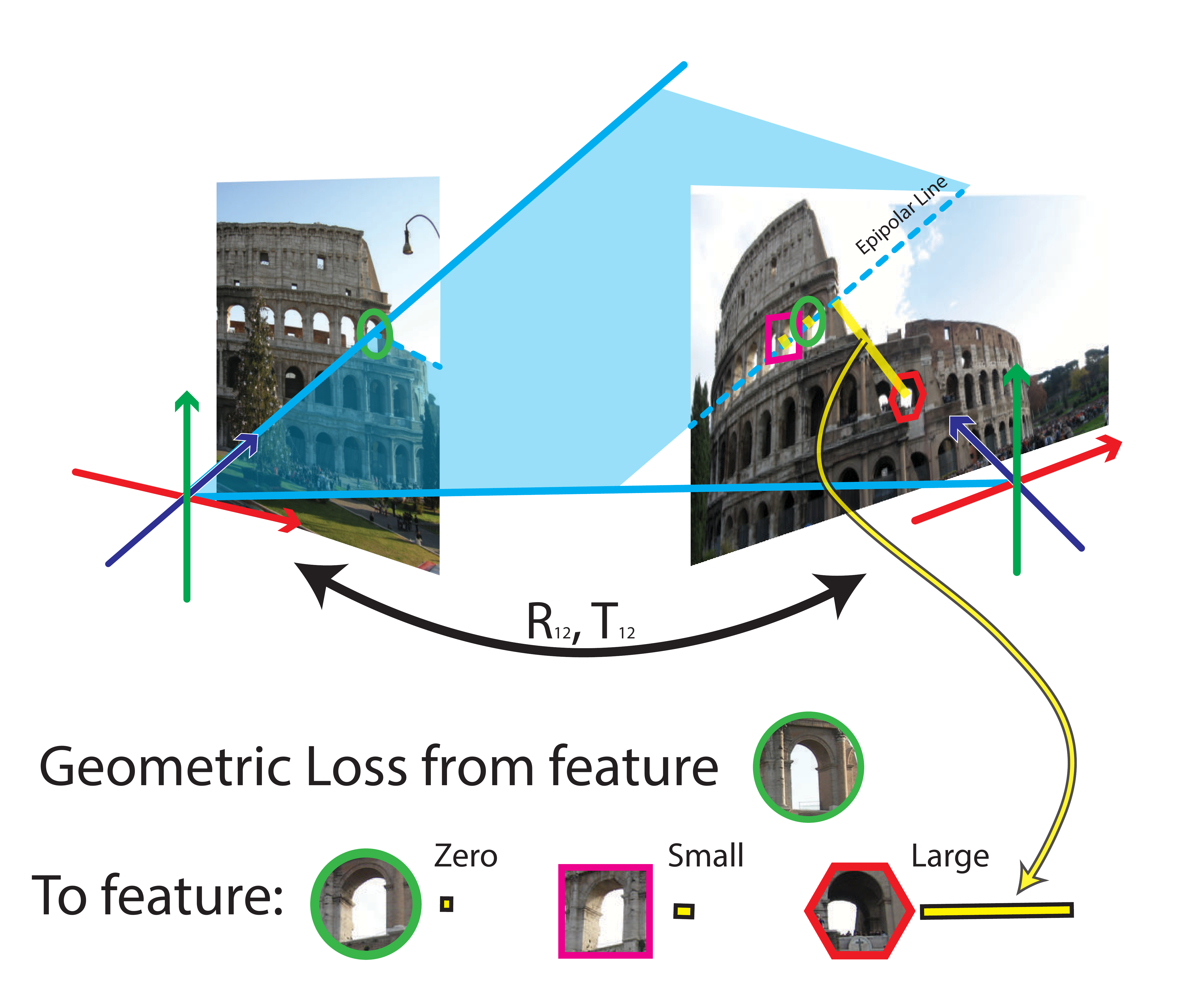}
\end{center}
  \caption{
    Illustrated here is an example of how the geometric loss is computed for one feature.  
    Errors are computed via absolute distance from the epipolar line, as expressed by (\ref{eq:essential_constraint}) via the epipolar constraint.
    The epipolar line is the line of projection of the feature on the first image, projected onto to the second.
    The distance to this line on the second image indicates how likely that point is to correspond geometrically to the original feature.
    There can be false positives along the projected line, as shown by the square feature in the figure, but other points will be eliminated, such as the hexagonal feature.
    Best viewed in color.
  }
\label{fig:geoconsist}
\label{fig:onecol}
\end{figure}

\subsection{Cycle Consistency}

Let $\mat{M}$ be the noiseless set of matches between our features, with $\mat{M}_{ij}$ being the matches between image $i$ and image $j$.
If the pairwise matches are globally consistent, then we can say that, for all $i, j, k$:
\begin{equation}
\mat{M}_{ij} = \mat{M}_{ik} \mat{M}_{kj}
\label{eq:cycconsist1}
\end{equation}
In other words, the matches between two images stay the same no matter what path is taken to get there. 
This constraint is known as \textit{cycle consistency}, and has been used in a number of works to optimize for global consistency \cite{zhou2015multi, wang2017multi, leonardos2016distributed}.
Stated in this form, there are $O(n^3)$ cycle consistency constraints to check.
A more elegant way to represent cycle consistency is to first create a `universe' of features that all images match to (see figure \ref{fig:universefeatures}).
Then, one can match the $i^{th}$ set of features to the universe using a matrix $\mat{X}_i$.
Then the cycle consistency becomes:
\begin{equation}
\mat{M}_{ij} = \mat{X}_{i}\mat{X}_{j}^\top
\label{eq:cycconsist2}
\end{equation}

\begin{table*}
\begin{center}
\begin{tabular}{|l|c|c|}
\hline
Method & Same Point Similarities & Different Point Similarities  \\
\hline\hline\hline
Ideal                              & 1.00e+0 $\pm$ 0.00e+0 & 0.00e+0 $\pm$ 0.00e+0 \\ \hline
Initialization Baseline            & 5.11e-1 $\pm$ 1.68e-2 & 2.56e-1 $\pm$ 2.06e-1 \\ \hline
3 Views, Noiseless                 & 9.96e-1 $\pm$ 7.70e-3 & 1.16e-1 $\pm$ 1.32e-1 \\ 
5 Views, Noiseless                 & 1.00e+0 $\pm$ 4.15e-4 & 1.22e-1 $\pm$ 1.67e-1 \\ \hline
3 Views, Added Noise               & 9.96e-1 $\pm$ 7.70e-3 & 1.16e-1 $\pm$ 1.32e-1 \\ 
5 Views, Added Noise               & 9.89e-1 $\pm$ 2.47e-2 & 7.67e-2 $\pm$ 1.56e-1 \\ 
6 Views, Added Noise               & 9.84e-1 $\pm$ 3.16e-2 & 7.46e-2 $\pm$ 1.57e-1 \\ \hline
3 Views, 5\% Outliers              & 9.29e-1 $\pm$ 1.79e-1 & 1.41e-1 $\pm$ 1.48e-1 \\ 
3 Views, 10\% Outliers             & 9.27e-1 $\pm$ 1.79e-1 & 1.40e-1 $\pm$ 1.51e-1 \\ \hline

\hline
\end{tabular}
\end{center}
\caption{
Results on Synthetic correspondence graphs.
The `Same Point Similarities' column is the mean and standard deviation of similarities for true corresponding points, while the `Different Point Similarities' is the same for points that do not correspond.
For the `Same Point Similarities' column higher is better, and for `Different Point Similarities' lower is better.
Losses tested against ground truth correspondence graph adjacency matrices.
Our method was not trained on ground truth correspondences but using unsupervised methods.
}
\label{fig:synthtable}
\end{table*}

\begin{figure}[t]
\begin{center}
  \includegraphics[width=0.8\linewidth]{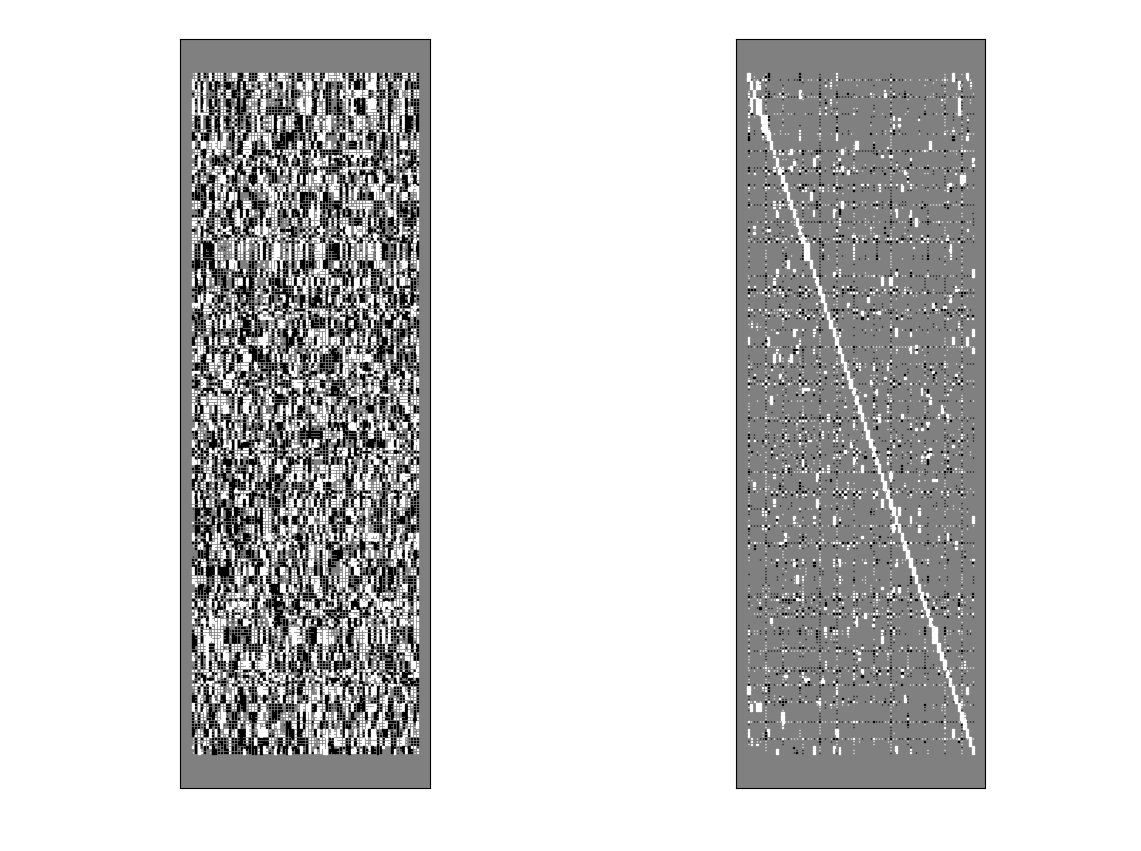}
  \end{center}
     \caption{
         Visualization of the learned embeddings using Hinton diagrams \cite{bremner1994hinton}.
         On the left we have the raw outputs, which are difficult to interpret.
         On the right, we rotated the features to best match the ground truth for a more interpretable visualization.
     }
  \label{fig:onecol}
  \label{fig:embeddingsviz}
\end{figure}

This reduces our complexity from $O(n^3)$ to $O(n^2)$.
We try to learn $\mat{E}$ to approximate $\mat{X}$ - in other words the final embedding should be an encoding of the universe of features. As we do not have the ground truth matches $\mat{M}$, we approximate it using the noisy adjacency matrix $\mat{A}$ of our correspondence graph. Thus our loss would be 
\begin{equation}
\mathcal{L}(\mat{A}, \mat{E}_n) = \mathcal{D}(\mat{A}, \mat{E}_n\mat{E}_n^\top)
\end{equation}
Here $\mathcal{D}$ could be an $L_2$ loss, $L_1$ loss, or many others. In this work, we use the $L_1$ loss. 
Note that because of this formulation, we can determine our embeddings only up to a rotation, as
\begin{equation}
\mat{E}R(\mat{E}R)^\top
= \mat{E}RR^\top\mat{E}^\top
= \mat{E}\mat{E}^\top
\label{eq:rotinvar}
\end{equation}
Thus when visualizing embeddings, we rotate them to make them more interpretable (see figure \ref{fig:embeddingsviz}).

\subsection{Geometric Consistency Loss}

One of the main advantages of this approach over more traditional optimization based approaches is the ability to add geometric consistency information into the loss at training time, even if it is not available at test time.
The simplest way to add geometric consistency losses, and the approach we use here, is to use the epipolar constraint.
The epipolar constraint describes how the positions of features in different images corresponding to the same point should be related.
Given a relative pose $(R_{ij}, T_{ij})$ between two cameras $i$ and $j$  (transforms $j$ to $i$) the epipolar on corresponding feature locations $X_i$ and $X_j$:
\begin{equation}
X_{i}^\top \cross{T_{ij}}R_{ij} X_{j} = 0
\label{eq:essential_constraint_rel}
\end{equation}
In this work we use the two pose epipolar constraint \cite{tron2014quotient}:
\begin{equation}
X_{i}^\top R_{i}^\top \cross{T_{j} - T_{i}}R_{j} X_{j} = 0
\label{eq:essential_constraint}
\end{equation}
The constraint assumes that the $X_k$ are calibrated i.e. the camera intrinsics are known. 
Given the matrix of ground truth correspondences $\mat{M}_{ij}$ between camera $i$ and $j$, we can formulate this as a loss:
\begin{equation}
\mathcal{L}_{ij,geom}(\mat{M}_{ij}) = \sum_{k,l} (\mat{M}_{ij})_{kl} \left|X_{i,k}^\top R_{i}^\top \cross{T_{j} - T_{i}}R_{j} X_{j,l}\right|
\label{eq:geom_cost}
\end{equation}
For our purposes, since we use low rank embeddings $\mat{E}_{i}$, $\mat{E}_{j}$, the loss would read:
\begin{align}
\mathcal{L}_{geom}(\mat{E})
=&\; \mathrm{tr}(\mat{G}^\top \mat{E}\mat{E}^\top) = \sum_{i,j} (\mat{E})_{i} \cdot (\mat{E})_{j} (\mat{G})_{ij} \\
(\mat{G})_{ij} =&\; \left|X_{i}^\top R_{c(i)}^\top \cross{T_{c(j)} - T_{c(i)}}R_{c(j)} X_{j}\right| \nonumber
\label{eq:geom_cost2}
\end{align}
Where $c(i)$ is the appropriate camera for point index $i$.

\begin{table*}
\begin{center}
\begin{tabular}{|l|c|c|c|}
\hline
Method (3 Views)                                          & $L_1$ Loss                          & $L_2$ Loss                          & Run Time (sec)                      \\
\hline\hline
MatchALS \cite{zhou2015multi} 15 Iterations               & 0.052 $\pm$ 0.003                   & 0.010 $\pm$ 0.002                   & 0.021 $\pm$ 0.003                   \\
MatchALS \cite{zhou2015multi} 25 Iterations               & 0.045 $\pm$ 0.007                   & 0.009 $\pm$ 0.003                   & 0.034 $\pm$ 0.003                   \\
MatchALS \cite{zhou2015multi} 50 Iterations               & 0.016 $\pm$ 0.008                   & 0.007 $\pm$ 0.003                   & 0.065 $\pm$ 0.006                   \\ \hline
PGDDS \cite{leonardos2016distributed} 15 Iterations       & 0.016 $\pm$ 0.002                   & 0.006 $\pm$ 0.002                   & 0.287 $\pm$ 0.043                   \\
PGDDS \cite{leonardos2016distributed} 25 Iterations       & 0.014 $\pm$ 0.002                   & 0.005 $\pm$ 0.002                   & 0.613 $\pm$ 0.089                   \\
PGDDS \cite{leonardos2016distributed} 50 Iterations       & 0.013 $\pm$ 0.002                   & 0.005 $\pm$ 0.002                   & 1.430 $\pm$ 0.234                   \\ \hline
Spectral                                                  & 0.054 $\pm$ 0.005                   & 0.018 $\pm$ 0.004                   & 0.018 $\pm$ 0.004                   \\ \hline 
\textbf{GCN, 12 Layers (ours)}                            & \textbf{0.025} $\pm$ \textbf{0.003} & \textbf{0.016} $\pm$ \textbf{0.003} & \textbf{0.039} $\pm$ \textbf{0.009} \\
\hline\hline
Method (4 Views)                                                         & $L_1$ Loss                          & $L_2$ Loss                          & Run Time (sec)                     \\
\hline\hline
MatchALS \cite{zhou2015multi} 15 Iterations                              & 0.064 $\pm$ 0.005                   & 0.012 $\pm$ 0.002                   & 0.030 $\pm$ 0.004                   \\
MatchALS \cite{zhou2015multi} 25 Iterations                              & 0.041 $\pm$ 0.010                   & 0.008 $\pm$ 0.004                   & 0.048 $\pm$ 0.005                   \\
MatchALS \cite{zhou2015multi} 50 Iterations                              & 0.011 $\pm$ 0.008                   & 0.005 $\pm$ 0.003                   & 0.094 $\pm$ 0.008                   \\ \hline
PGDDS \cite{leonardos2016distributed} 15 Iterations                      & 0.015 $\pm$ 0.002                   & 0.006 $\pm$ 0.001                   & 0.436 $\pm$ 0.090                   \\
PGDDS \cite{leonardos2016distributed} 25 Iterations                      & 0.014 $\pm$ 0.002                   & 0.005 $\pm$ 0.001                   & 0.961 $\pm$ 0.181                   \\
PGDDS \cite{leonardos2016distributed} 50 Iterations                      & 0.013 $\pm$ 0.002                   & 0.005 $\pm$ 0.002                   & 2.056 $\pm$ 0.424                   \\ \hline
Spectral Method                                                          & 0.055 $\pm$ 0.004                   & 0.017 $\pm$ 0.003                   & 0.028 $\pm$ 0.003                   \\ \hline
\textbf{GCN, 12 Layers (ours)}                                           & \textbf{0.023} $\pm$ \textbf{0.003} & \textbf{0.015} $\pm$ \textbf{0.002} & \textbf{0.056} $\pm$ \textbf{0.017} \\ \hline
\end{tabular}
\end{center}

\caption{
Results on Rome16K Correspondence graphs, showing the mean and standard deviation of the $L_1$ and $L_2$.
Our method was not trained on ground truth correspondences but using unsupervised methods and geometric side losses.
As our method gives soft labels, we use cannot use precision or recall as is standard in testing cycle consistency \cite{zhou2015multi}.
Thus we test against ground truth correspondence graph adjacency matrices computed from the bundle adjustment output.
Our method performs better than 25 iteration of the MatchALS \cite{zhou2015multi} method, but does not perform as well as 50 iterations.
We do not perform as well as the Projected Gradient Descent Doubly Stochastic (PGDDS) \cite{leonardos2016distributed} but we perform significantly faster than them.
Note that we perform much better in $L_1$ performance rather than $L_2$, as we optimized the network weights using an $L_1$ loss.
}
\end{table*}

\section{Experiments}

\begin{figure*}
\begin{center}
  \includegraphics[width=0.8\linewidth]{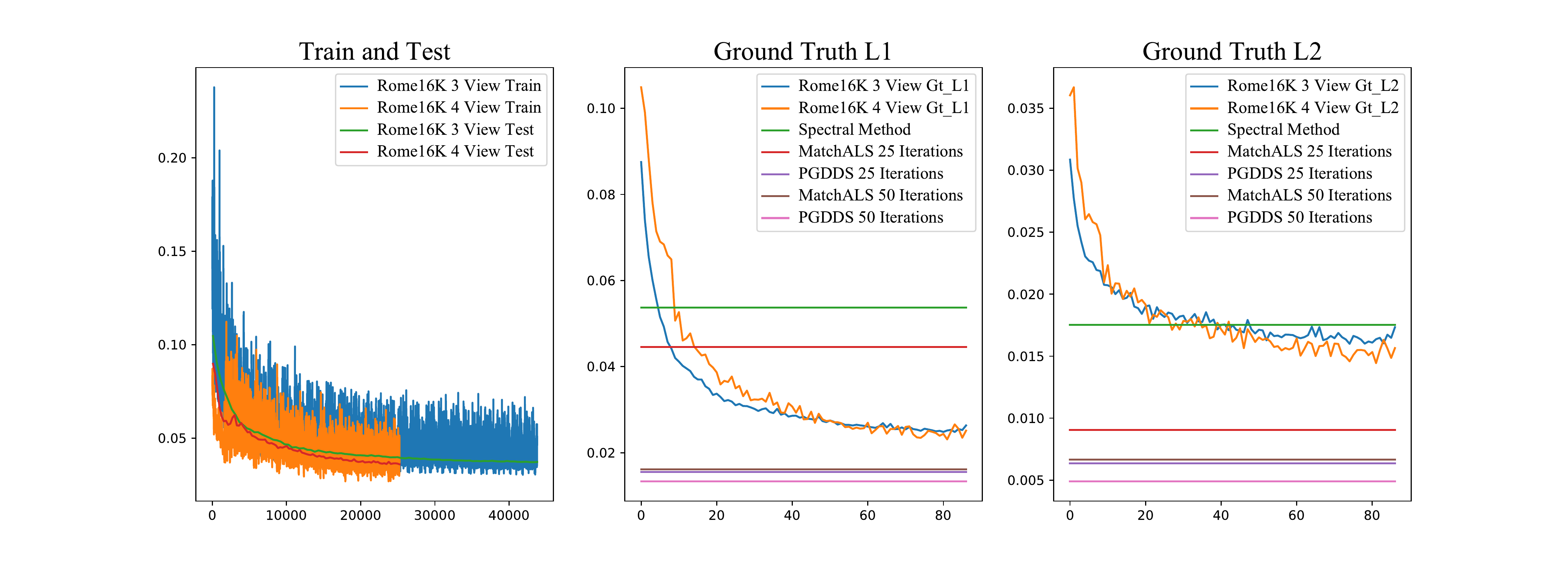}
  \end{center}
    \caption{
      Training curves with and without Geometric Training loss.
      The geometric training loss improves testing performance.
      Shown as horizontal lines are the state of the art optimization based baselines.
      Even with a simple network we compare well with them.
    }
  \label{fig:short}
\end{figure*}

\begin{figure*}
\begin{center}
  \includegraphics[width=0.8\linewidth]{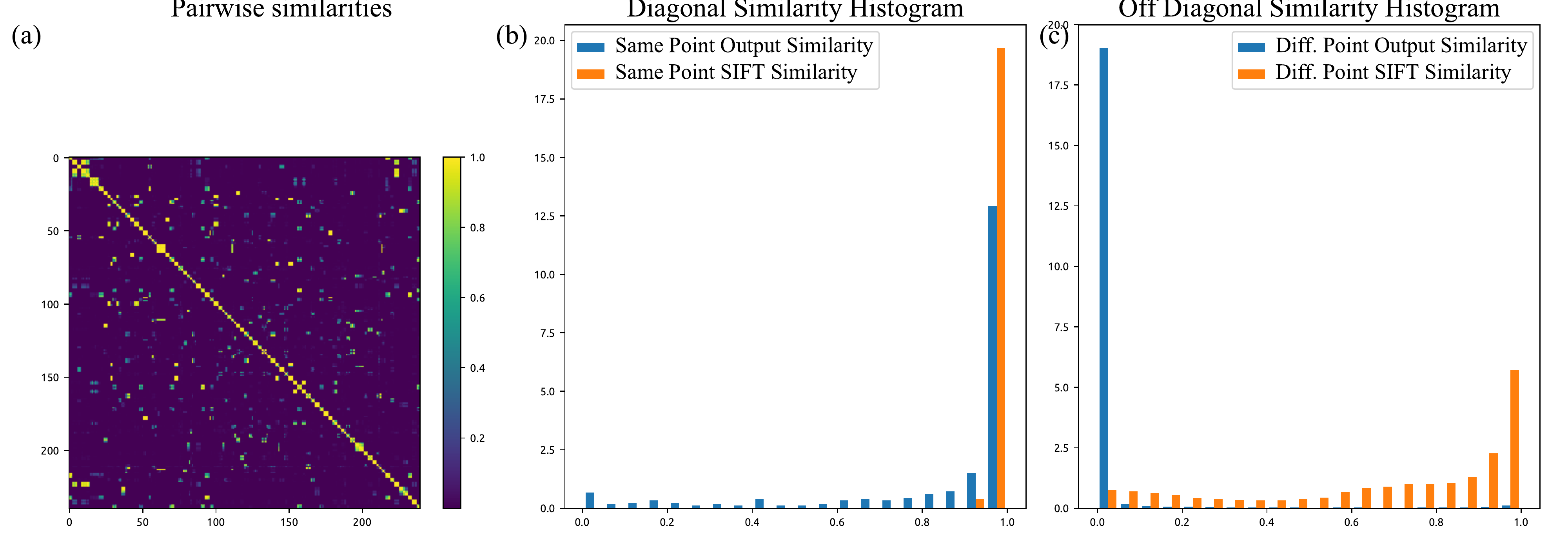}
  \end{center}
     \caption{Example output of our network. (a) Similarity Matrix of the of the embeddings (b) Histogram of feature similarities for pairs which correspond to the same 3D point, comparing our output features with the original SIFT features (c) Histogram of feature similarities for pairs which correspond to different 3D points, comparing our output features with the original SIFT features}
  \label{fig:short}
\end{figure*}
\subsection{Synthetic Graph Dataset}
We first test our method on synthetically generated data as a simple proof of concept.
To generate the data, we generate $p$ points, each with its own randomly generated descriptor.
To create the graph, we generate random permutation matrices, with a noise applied to it after it is generated.
We initialize the input descriptors using the true descriptor, plus some added Gaussian noise.
No geometric losses were added during training for these experiments.
However, the method was robust in testing with different noise functions and parameters.
The normalized noisy input descriptors are our baseline - they correlate with the true values but do not preserve the structure well.
However, the GCN recovered the true structure very well, as shown in Table \ref{fig:synthtable}.
All experiments were run with a 12 layer GCN with the ReLU nonlinearity and skip connections.
All were trained with the Adam optimizer \cite{kingma2014adam} and a learning rate $10^{-4}$

\subsection{Rome 16K Graph Dataset}
We  use the Rome16K dataset \cite{li2010location} to test our algorithm in real world settings.
Rome16K consists of 16 thousand images of various historical sites in Rome extracted from Flickr, along with the 3D structure of the sites provided by bundle adjustment.
While not a standard dataset to test cycle consistency, other datasets had insufficient data to train a network on.
Rome16K is typically used to test bundle adjustment methods.
Therefore, to use our method, we extract image triplets and quadruplets with overlap of 80 points or more to test our algorithm, with the points established as corresponding in the given bundle adjustment output.
For the initial embedding we use the original 128 dimensional SIFT descriptors, normalized to have unit $L_2$ norm, the calibrated x-y position, the orientation, and log scale of the SIFT feature.

For these experiments we train with the $L_1$ norm and geometric consistency losses.
We evaluate on a test set using the ground truth adjacency matrix, which we compute from the bundle adjustment given by the Rome16K dataset.
Traditional methods of evaluation use Precision and Recall, after making the labels 0 or 1.
We forgo this as our method was trained to output soft labels - we in turn make the labels soft for the optimization based methods.
For this method to work, we need the dimension of the embedding to be at least the number of unique points in the scene.
Picking the correct number is difficult a-priori, and is a problem with all cycle consistency based methods.
Here we use the ground truth dimension of the embedding to test both our method and the baselines.
For implementation details, our network has 12 layers, and was trained with the Adam optimizer \cite{kingma2014adam} with a learning rate of $10^{-4}$, with an exponentially decaying learning rate. We incorporate skip connections between the input, $6^{th}$, and $12^{th}$ layers, which greatly improved training convergence and testing performance (see supplemental material). 

We compare our method to spectral and optimization based baselines with different maximum iteration cutoffs.
Our method is signification better than a simple spectral based optimization.
Our network, though only 12 layers, has comparable accuracy to MatchALS \cite{zhou2015multi} run between 25 and 50 iterations.
Although our method does not outperform the Projected Gradient Descent - Doubly Stochastic (PGDDS) \cite{leonardos2016distributed} method, we have significantly better run time.
Additional comparisons to different iterations numbers of iterations can be found in the supplementary material.

\section{Conclusion}

We have shown a novel method for training feature matching using GCNs, using an unsupervised cycle consistency loss and geometric consistency losses.
We have demonstrated its effectiveness on the Rome16K dataset compared to the traditional optimization based baselines on a simple GCN.
For future work, we will investigate robust losses for better outlier rejection, and using higher order geometric constraints, such as the tri-focal tensor, as additional loss terms.
A long term goal is to incorporate learning image level features into this pipeline.
Moreover we hope to adapt this method to a distributed setting, which would be a simple extention given current approach.

\subsubsection*{Acknowledgements}
Support by ARL DCIST CRA W911NF-17-2-0181, NSF-IIS-1703319, ARL RCTA W911NF-10-2-0016, ONR N00014-17-1-2093, and the Honda Research Institute is gratefully acknowledged.

{\small
\bibliographystyle{ieee}
\bibliography{mybib}
}

\newpage
\onecolumn

\section*{Appendix}

\subsection*{A. Relative Efficiency Against Optimization Based Methods}

\begin{figure*}[ht]
\begin{center}
  \includegraphics[width=0.9\linewidth]{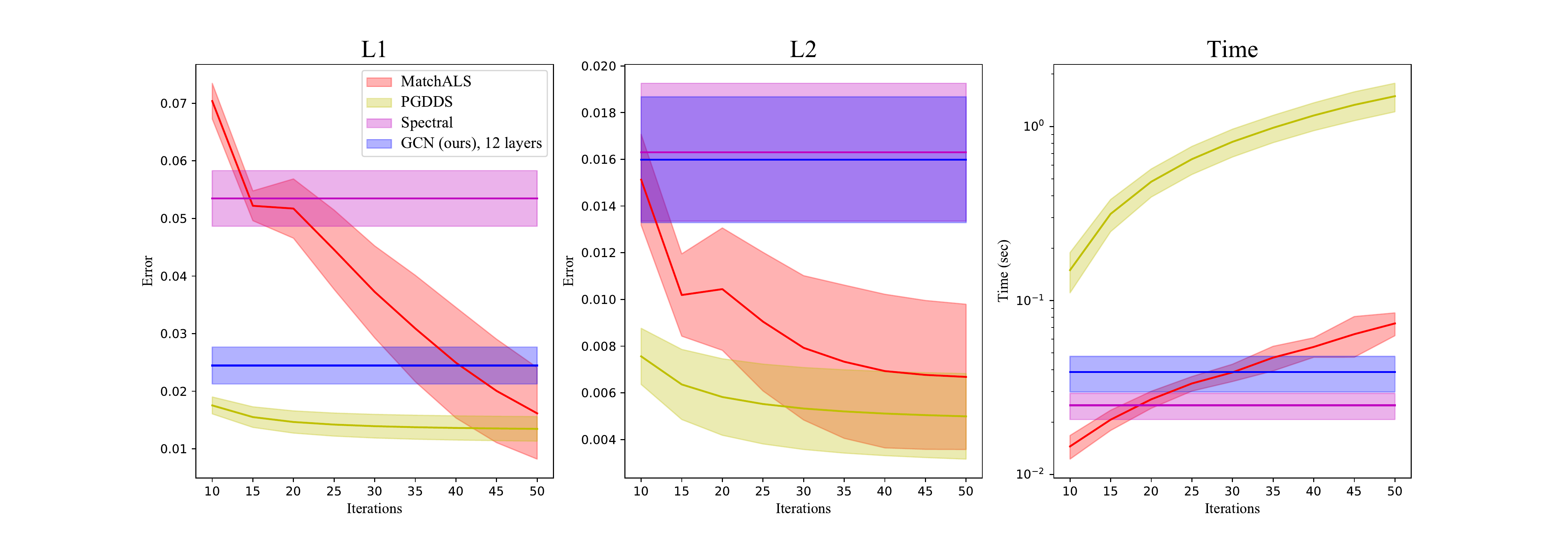}
\end{center}
  \caption{
    This figure plots the average accuracy of the optimization methods by iteration compared to 12 iterations of our GCN based method.
    The bold center line shows the mean, the colored in region shows 1 standard devation of the errors.
    As we optimize for $L_1$ error, our method's $L_1$ loss is substantially better than its $L_2$ losses.
    The right most plot shows time, plotted on a log scale for clarity.
  }
\label{fig:iterplot}
\label{fig:onecol}
\end{figure*}

In figure \ref{fig:iterplot} we show a more fine grained comparison of our algorithm to optimization based methods at different numbers of iterations.
We show the mean and standard devation of the errors of all methods, testing the optimization based methods at different iterations.

\subsection*{B. Effect of Group Normalization}
\begin{figure*}[!hb]
\begin{center}
  \includegraphics[width=0.8\linewidth]{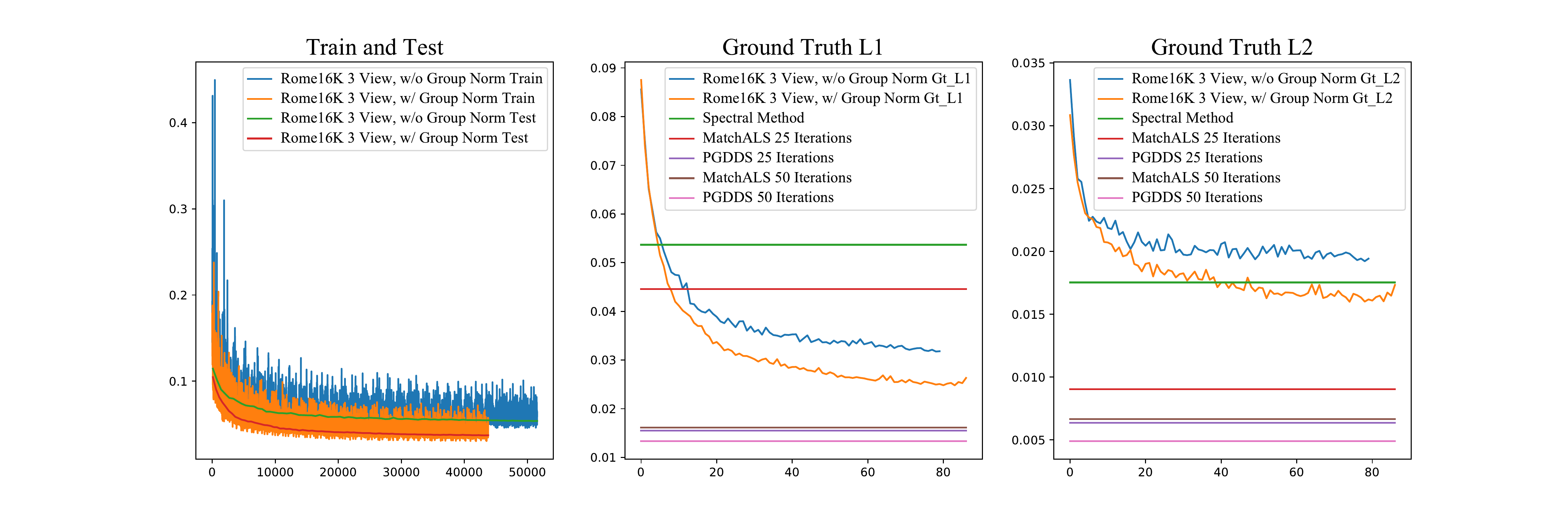}
\end{center}
  \caption{
    This figure plots the training curves of the network with and without Group Normalization.
    Note that while the train/test curves are fairly similar with or without Group Normalization, distance to the ground truth is improved greatly with Group Normalization.
  }
\label{fig:groupnorm}
\label{fig:onecol}
\end{figure*}

We observe that training with group normalization helps with training and generalization error, as shown in figure \ref{fig:groupnorm}.
As the Group Normalization operation is not easily parallelized across nodes in the graph, the use of Group Normalization is the main barrier to parallelization.
Thus we will need to find alternative more parallelizable methods to improve generalization error for future work.

\end{document}